\documentclass[lettersize,journal]{IEEEtran}
\usepackage{amsmath,amsfonts}
\usepackage{algorithmic}
\usepackage{algorithm}
\usepackage{array}
\usepackage{subcaption}
\usepackage{textcomp}
\usepackage{stfloats}
\usepackage{url}
\usepackage{verbatim}
\usepackage{graphicx}
\usepackage{cite}
\usepackage{hyperref}
\RequirePackage{xspace}
\usepackage{marvosym}
\newcommand{\momo}{{\sc Herman}}
\usepackage[dvipsnames]{xcolor}
\definecolor{cvprblue}{rgb}{0.21,0.49,0.74}
\definecolor{LightCyan}{rgb}{0.88,1,1}
\definecolor{LightPurple}{HTML}{D6D6FF}
\usepackage{multirow}
\usepackage{marvosym}

\usepackage{colortbl}
\usepackage{amssymb}
\usepackage{amsmath}

\newcommand{\eg}{\textit{e.g.\xspace}}
\usepackage{booktabs}
\usepackage{bm}
\usepackage{arydshln}

\usepackage{amsmath}    % 数学公式支持
\usepackage{listings}   % 代码格式化
\usepackage{xcolor}     % 颜色支持
\usepackage{fancyvrb}   % 代码高亮

\begin{document}

\title{\momo: Hierarchical Representation Matching for \\CLIP-based Class-Incremental Learning}

\author{Zhen-Hao Xie, Yan Wang, Lan Li, Han-Jia Ye, De-Chuan Zhan, Da-Wei Zhou
\thanks{Zhen-Hao Xie, Yan Wang, Lan Li, Han-Jia Ye, De-Chuan Zhan and Da-Wei Zhou are with the School of Artificial Intelligence and the State Key Laboratory for Novel Software Technology, Nanjing University, Nanjing, China (e-mail: wenzh@lamda.nju.edu.cn, wangy@lamda.nju.edu.cn, lil@lamda.nju.edu.cn, yehj@lamda.nju.edu.cn, zhandc@lamda.nju.edu.cn, zhoudw@lamda.nju.edu.cn).
}% <-this % stops a space
\thanks{Corresponding author: Da-Wei Zhou}}

\markboth{Journal of \LaTeX\ Class Files,~Vol.~14, No.~8, August~2021}%
{Shell \MakeLowercase{\textit{et al.}}: A Sample Article Using IEEEtran.cls for IEEE Journals}


\maketitle

\begin{abstract}
Class-Incremental Learning (CIL) aims to acquire new classes while preserving prior knowledge. Pre-trained models like CLIP provide a powerful foundation with their cross-modal alignment and multi-level visual representations. However, existing CLIP-based CIL methods typically rely on simplistic templates and last-layer alignment, overlooking the hierarchical nature of visual concepts. This makes knowledge brittle, as updates for fine-grained distinctions are forced into the same space encoding coarser separations. For instance, after learning ``cat" and ``dog", adding ``wolf" requires encoding subtle differences (\eg, ear shape). Modeled at a single level, these updates can inadvertently shift the boundary separating canines from felines, degrading performance on old classes.
To address this, we propose HiErarchical Representation MAtchiNg (\momo). \momo~leverages LLMs to generate hierarchical textual descriptors and aligns them with CLIP's intermediate visual layers, creating a structured semantic space. An adaptive routing mechanism then dynamically balances contributions from different levels, allowing coarse layers to stabilize high-level separation while fine layers handle subtle distinctions. To ensure continual adaptation, the router is updated via a projection-based strategy that preserves learned subspaces for old tasks while capturing new information in complementary directions. By structuring and adaptively routing hierarchical semantics, \momo~enhances discrimination and mitigates forgetting, achieving SOTA results on standard CIL benchmarks.
\end{abstract}


\section{Introduction}
Deep learning has achieved remarkable progress across domains~\cite{Hinton06, deng2009imagenet, lecun2015deep, he2015residual, goodfellow2016deep}. Yet real-world data arrive as streams that demand continuous adaptation. Class-Incremental Learning (CIL) addresses this by gradually incorporating new categories~\cite{zhao2020maintaining, masana2022class, DBLP:conf/cvpr/YuZ0H0LH24, dohare2024loss, wang2024comprehensive, lai2025order}, but a major obstacle is catastrophic forgetting~\cite{serra2018overcoming, ramasesh2021effect, shi2021overcoming, 11275914}, where learning new classes diminishes previously acquired knowledge of old ones. While early approaches often train models from scratch~\cite{DBLP:journals/pami/LiH18a, rebuffi2017icarl, DBLP:conf/cvpr/YanX021}, recent work leverages pre-trained models (PTMs) such as CLIP~\cite{radford2021learning}. Pre-trained on vast image–text data~\cite{DBLP:journals/corr/abs-2111-02114}, CLIP provides rich prior knowledge and strong cross-modal alignment, offering a solid foundation for continual adaptation~\cite{wang2022learning, DBLP:conf/eccv/0002ZESZLRSPDP22, DBLP:conf/cvpr/SmithKGCKAPFK23, huang2024class}.

However, existing CLIP-based CIL methods have not fully exploited CLIP’s capabilities. Most simply match the final-layer visual representation to a textual embedding derived from a fixed template, \eg, ``a photo of a \texttt{[CLASS]}''. This design has two limitations. First, single-level, coarse-grained alignment makes knowledge brittle: fine-grained updates are forced into the same space that encodes coarse separations. For example, after a model learns to separate ``cat'' and ``dog'', incrementally adding ``wolf'' requires encoding subtle cues (\eg, snout shape, ear posture). When modeled in a single space, these updates can inadvertently shift the decision boundary that separates canines from felines, degrading performance on old classes. Second, this approach underutilizes CLIP’s inherent capacity for hierarchical semantics. Pre-training on large-scale image–text corpora exposes CLIP to captions that often include fine-grained attributes~\cite{DBLP:journals/corr/abs-2111-02114, 11360301}, and its vision encoder thus captures multi-level features, and its text encoder can represent descriptors at varying levels of abstraction. But most methods focus solely on final-layer alignment.

Leveraging multi-level features offers a direct remedy to brittle knowledge. A hierarchical representation space can decouple learning across semantic granularities: coarse-grained representations (\eg, ``animal'', ``canine'') can anchor separations among distant categories such as ``cat'' and ``dog'', while fine-grained representations (\eg, ``pointed ears'', ``bushy tail'') can resolve subtle distinctions between ``dog'' and ``wolf''. Crucially, updates needed for fine-grained distinctions can be concentrated at finer levels of the hierarchy, leaving robust coarse-grained anchors largely undisturbed. This isolates the acquisition of new details from foundational concepts, preventing destructive boundary shifts and thus mitigating catastrophic forgetting.

However, introducing hierarchy raises two challenges. First, how should contributions from different hierarchical levels be integrated? The importance of each level is input- and task-dependent: separating ``car'' from ``bicycle'' may require only coarse shape cues, whereas distinguishing visually similar bird species may hinge on fine-grained color or pattern details~\cite{wu2024llmclip, dai2025propvg}. Static or uniform averaging is therefore suboptimal. Second, in the incremental setting, how can the integration mechanism itself adapt without forgetting? A router that learns to prioritize fine-grained cues for a new task may forget the routing policy that favored coarser cues for older tasks, reintroducing performance degradation.

To address these challenges, we propose HiErarchical Representation MAtchiNg (\momo), a CLIP-based CIL framework that structures, aligns, and adaptively routes hierarchical semantics. First, \momo\ uses large language models (LLM)~\cite{OpenAI2025GPT5, qwen3} to generate multi-level textual descriptors for each class and explicitly aligns them with intermediate visual layers, forming a structured semantic space. For each layer, we select the top-K most aligned descriptors and aggregate them into a compact textual embedding, providing supervision beyond the final layer. Second, an input-conditioned adaptive router dynamically weighs hierarchical levels per instance, prioritizing the most informative cues for discrimination. Third, to enable continual adaptation while curbing forgetting, we update the router via a projection-based strategy that preserves informative subspaces from old tasks and absorbs new knowledge in orthogonal directions. We further train with a lightweight visual adapter and a contrastive objective that mixes hierarchical descriptors with class templates, and employ feature-level generative replay~\cite{DBLP:conf/cvpr/ZhuZWYL21} to stabilize past knowledge without storing raw images. Together, these components enhance discrimination, effectively mitigate catastrophic forgetting, and yield state-of-the-art performance on standard CIL benchmarks.

\section{Related Work}
\label{Related Work}
\paragraph{Pre-Trained Model-Based CIL}
Early approaches to class-incremental learning (CIL)~\cite{rebuffi2017icarl, DBLP:journals/pami/LiH18a, de2021survey, DBLP:conf/cvpr/YanX021, masana2022class} typically trained models from scratch, which often led to limited generalization and severe forgetting when adapting to new tasks. To overcome these drawbacks, recent studies have shifted towards leveraging large-scale pre-trained models (PTMs), whose rich prior knowledge provides a stronger foundation for continual learning~\cite{wang2024model, yeongbin2024trainattention, Yu_2025_CVPR, ven2025on, 11249423}. A common strategy is to freeze the pre-trained backbone and introduce lightweight modules, such as prompts~\cite{DBLP:conf/eccv/0002ZESZLRSPDP22, wang2022learning, wang2022s, zhou2022learning, DBLP:conf/cvpr/SmithKGCKAPFK23} and adapters~\cite{rebuffi2017learning, chen2022adaptformer, DBLP:conf/cvpr/YuZ0H0LH24}. For example, L2P~\cite{wang2022learning} and DualPrompt~\cite{DBLP:conf/eccv/0002ZESZLRSPDP22} maintain a pool of visual prompts~\cite{jia2022visual} and dynamically select instance-specific prompts for each input when fine-tuning a pre-trained Vision Transformer. Other methods design more sophisticated prompt strategies, such as combining or generating prompts through attention mechanisms~\cite{DBLP:conf/cvpr/SmithKGCKAPFK23, wang2023hide, DBLP:conf/aaai/LiZ25a} or generative networks~\cite{jung2023generating, 11329196}. Prototype-based methods further exploit PTM features by directly constructing classifiers via class prototypes~\cite{snell2017prototypical, zhou2023revisiting, mcdonnell2023ranpac}. When CLIP is adopted as initialization, multi-modal prompts are explored to better align vision and language representations~\cite{wang2022s, 10.5555/3600270.3601550, wang2023attriclip, 11275895}. Beyond prompts, MOE-Adapter~\cite{DBLP:conf/cvpr/YuZ0H0LH24} extends lightweight module selection with a MoE design~\cite{masoudnia2014mixture}, while RAPF~\cite{huang2024class} enhances adapters through decomposed parameter fusion to mitigate forgetting.

\paragraph{Textual descriptors for CLIP}
PTMs, such as CLIP~\cite{radford2021learning}, align images and texts in a shared embedding space, achieving strong zero-shot recognition and retrieval. However, their performance in specialized domains, such as healthcare, remains limited due to the need for fine-grained visual cues and domain-specific semantics. To address this, a variety of lightweight adaptation strategies have been proposed, with text prompting attracting particular attention. Existing approaches can be broadly categorized into two main types: soft prompting and hard prompting. Soft prompting~\cite{zhou2022learning, wu2024controlmllm, nemesis, Tian_2024_CVPR, zhang2024adversarial} introduces learnable tokens into the textual input, enabling parameter-efficient adaptation while keeping the pre-trained backbone frozen. In contrast, hard prompting~\cite{NEURIPS2023_a0054803, yu2024api, 11341891} directly inserts natural language phrases into predefined templates, requiring no additional training and offering better interpretability. However, both soft and hard prompting still fall short in capturing the hierarchical representations that are crucial for CIL. Motivated by this limitation, we propose to extend hard prompting by introducing hierarchical textual descriptors. These descriptors serve as semantically rich anchors that transcend conventional flat templates, allowing models to more effectively capture and enhance recognition in hierarchical representation spaces.

\section{Preliminaries}
\subsection{Problem Setup: Class-Incremental Learning}
CIL aims to continually extend a unified classifier as novel classes arrive in a sequence of tasks~\cite{rebuffi2017icarl}. Let the training stream be $\{D_1, D_2, \ldots, D_T\}$, where each incremental task $D_t=\{(\mathbf{x}_i,y_i)\}_{i=1}^{n_t}$ contains $n_t$ labeled instances. Each input $\mathbf{x}_i \in \mathcal{X}$ belongs to a class $y_i \in Y_t$, where $Y_t$ is the label set of task $t$. We assume disjoint label spaces across tasks, \textit{i.e.}, $Y_t \cap Y_{t'}=\varnothing$ for $t\neq t'$. Following the \emph{exemplar-free} protocol~\cite{DBLP:conf/cvpr/ZhuZWYL21, DBLP:conf/eccv/0002ZESZLRSPDP22, wang2022learning}, the learner cannot store or replay samples from past tasks. When learning the $t$-th task, only $D_t$ is accessible. The objective is then to build a unified classifier over all classes observed so far, $\mathcal{Y}_t = Y_1 \cup \cdots \cup Y_t$, by finding a function $f: \mathcal{X} \rightarrow \mathcal{Y}_t$ that minimizes the empirical risk:
\begin{equation}
    f^\star = \arg\min_{f \in \mathcal{H}} \; \mathbb{E}_{(x,y)\sim D_1 \cup \cdots \cup D_t} \; \mathbb{I}\big( y \neq f(x) \big),
\end{equation}
where $\mathcal{H}$ denotes the hypothesis space and $\mathbb{I}(\cdot)$ is the indicator function.

\subsection{CLIP-Based Classification}
Following \cite{zhou2023learning,zhou2025external}, we assume a pre-trained CLIP is available for classification. Given an input image $\mathbf{x}\in\mathbb{R}^{H\times W\times C}$, the vision encoder $g_v(\cdot)$ first partitions it into a sequence of flattened 2D patches $\mathbf{x}_e \in \mathbb{R}^{(L-1)\times D}$, where $(L-1)$ denotes the number of patch tokens and $D$ is the embedding dimension. A learnable \verb|[CLS]| token $\mathbf{x}_{\text{cls}} \in \mathbb{R}^D$ is prepended to obtain the sequence $\mathbf{x}_p = [\mathbf{x}_{\text{cls}}; \mathbf{x}_e]$, which is processed by $B$ layers. The representation at the $b$-th layer is:
\begin{equation}
\label{eq:inter-visual-embeddings}
    \mathbf{x}^b = [\mathbf{x}_{\text{cls}}^b; \mathbf{x}_{1}^b; \dots; \mathbf{x}_{L-1}^b] \in \mathbb{R}^{L \times D},
\end{equation}
where $\mathbf{x}_{\text{cls}}^b \in \mathbb{R}^{D}$ is the class token and $\mathbf{x}_l^b$ is the patch token.

For each class $i \in \mathcal{Y}_t$, we utilize a templated prompt $\mathbf{c}_i$ (\textit{e.g.}, ``a photo of a [CLASS]'') and obtain its textual embedding $\mathbf{e}_i = g_t(\mathbf{c}_i)$ via the text encoder~\cite{radford2021learning}. Classification is performed by measuring the similarity between the final-layer \verb|[CLS]| token $\mathbf{x}_{\text{cls}}^B$ and these textual embeddings. The probability of class $i$ is then obtained as a Softmax over the cosine similarities:
\begin{align}
    \label{eq:clip-loss}
    \begin{aligned}
        f_{y_i}(\mathbf{x}) 
        &= \frac{\exp\!\big(\cos(\mathbf{x}_{\text{cls}}^B, \mathbf{e}_i)\big)}{\sum_{j=1}^{|\mathcal{Y}_t|} \exp\!\big(\cos(\mathbf{x}_{\text{cls}}^B, \mathbf{e}_j)\big)}\\
        &= \frac{\exp\!\big(\cos(g_v(\mathbf{x}), g_t(\mathbf{c}_i))\big)}{\sum_{j=1}^{|\mathcal{Y}_t|} \exp\!\big(\cos(g_v(\mathbf{x}), g_t(\mathbf{c}_j))\big)},
    \end{aligned}
\end{align}
where $\cos(\cdot, \cdot)$ denotes cosine similarity. In this formulation, CLIP performs classification by aligning visual embeddings with textual prompts, which can be seamlessly extended to CIL by enlarging the label space $\mathcal{Y}_t$ as new categories arrive~\cite{huang2024class}.

\noindent\textbf{Discussions.}
As shown in Eq.~\ref{eq:clip-loss}, classification depends only on the final-layer visual embedding aligned with a simple templated prompt. However,  the hierarchical representations generated by the vision encoder $g_v(\cdot)$ in Eq.~\ref{eq:inter-visual-embeddings} are overlooked. These intermediate embeddings encode hierarchical semantics that can further enhance the final representation. Likewise, textual representations should not be constrained to fixed templates; instead, they should also exhibit hierarchical structures. By incorporating hierarchical visual and textual representations, it becomes possible to perform cross-modal alignment in a hierarchical view. Furthermore, an effective mechanism is required to dynamically exploit such hierarchical information while continually updating representations without incurring catastrophic forgetting.

\section{Hierarchical Representation Matching}
\label{method}

\begin{figure*}
    \centering
    \includegraphics[width=\linewidth]{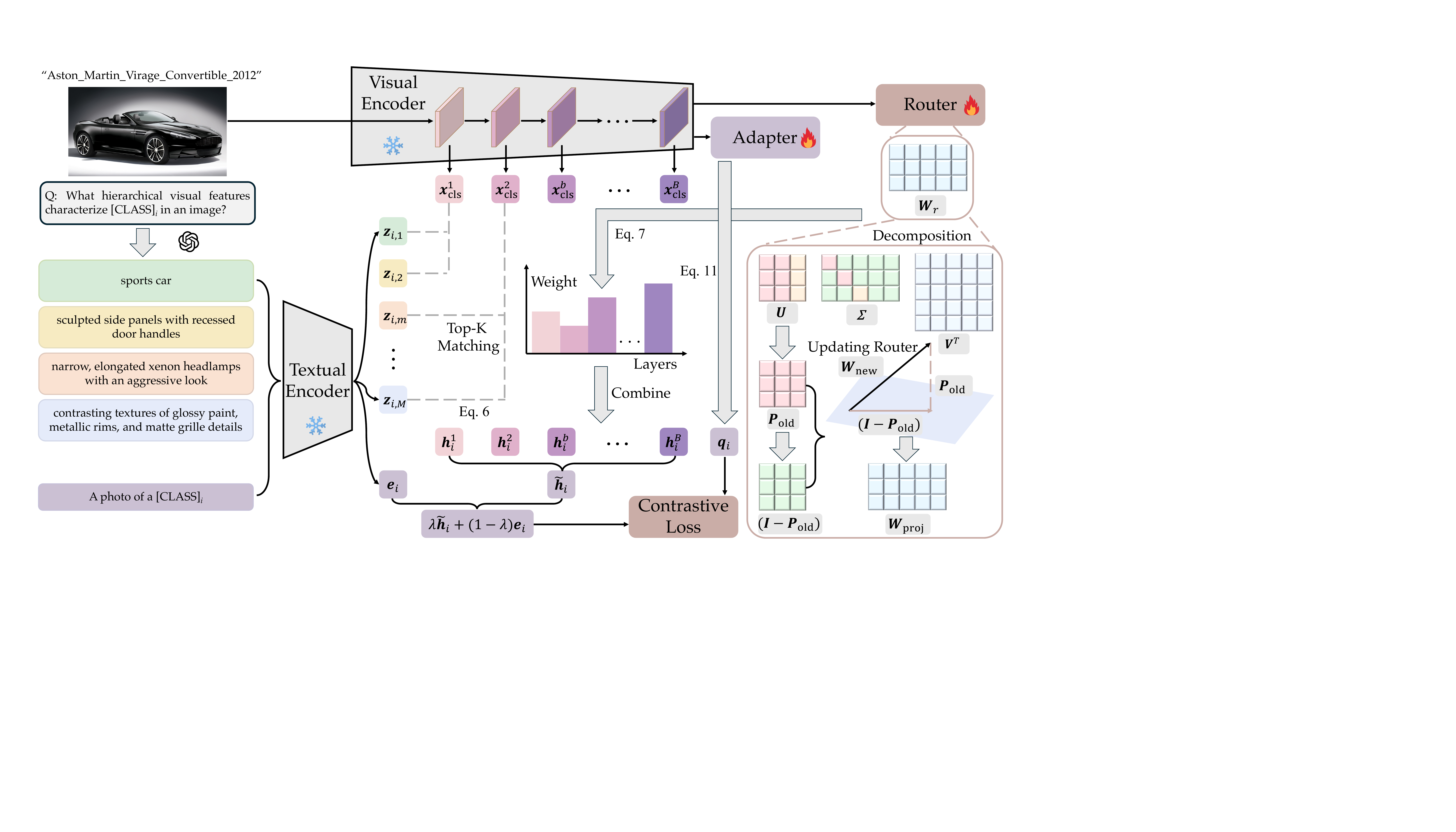}
    \caption{Illustration of \momo. The model leverages LLMs to generate hierarchical textual descriptors and explicitly matches them with intermediate visual features of CLIP. An adaptive routing mechanism then balances contributions across different hierarchical levels, while a projection-based update stabilizes routing to alleviate catastrophic forgetting in continual learning.}
    \label{fig:overview}
 \vspace{-5mm}
\end{figure*}
To address the observed challenges, we introduce hierarchical textual descriptors that mirror the multi-level structure encoded in intermediate visual representations. These descriptors are aligned with visual features at corresponding hierarchical levels, providing auxiliary supervision throughout the hierarchy and strengthening vision–language alignment. In addition, we further employ an adaptive routing mechanism that regulates the contributions of different hierarchical levels for each input. To stabilize routing and alleviate catastrophic forgetting, the router is updated through a projection that preserves informative subspaces of prior knowledge while enabling compact adaptation to new categories.

\subsection{Obtaining Hierarchical Representations}
In visual recognition, the semantic information required for discrimination is not uniform across categories. Coarse-grained descriptors, such as \textit{``animal''} versus \textit{``vehicle''}, are sufficient to separate semantically distant classes like \textit{``cat''} and \textit{``car''}. In contrast, distinguishing semantically similar categories such as \textit{``cat''} and \textit{``lion''} requires fine-grained descriptors, such as \textit{``soft and finely textured fur''} or \textit{``thick and elongated whiskers''}. Conventional CLIP, however, relies on a single fixed prompt aligned with the final-layer representation, overlooking such hierarchical cues. To address this, we introduce hierarchical textual descriptors that range from coarse to fine-grained levels, aligning them with intermediate visual features to provide richer supervision.

To capture semantic information at different levels of abstraction, we construct a set of natural language descriptors for each class that spans from coarse-grained cues to fine-grained details. Concretely, for each class $i \in \mathcal{Y}_t$, we \textit{recursively} generate a collection of $M$ descriptors $\mathcal{T}_i = \{\mathbf{t}_{i,1}, \dots, \mathbf{t}_{i,M}\}$ with the aid of LLMs. These descriptions are designed to progress from high-level category cues to increasingly detailed characteristics. To illustrate, we take the class \textit{``Aston\_Martin\_Virage\_Convertible\_2012''} from StanfordCars~\cite{krause20133d} as an example:
\begin{itemize}[label={}]
    \item\textbf{Q:} \textit{What hierarchical visual features characterize $\text{[CLASS]}_i$ in an image?}
    \item\textbf{A:} \textit{There are several useful visual features to tell there is a $\text{[CLASS]}_i$ in an image:}
    \begin{itemize}[label={--}]
        \item \textit{sports car}
        \item \textit{sculpted side panels with recessed door handles}
        \item \textit{contrasting textures of glossy paint, metallic rims, and matte grille details}
    \end{itemize}
\end{itemize}
At the coarsest level, the descriptors emphasize only the overall identity of the object, while intermediate ones highlight representative architectural elements, and the finest-grained descriptors capture rich contextual details by combining multiple attributes together. Once generated, these descriptors are mapped into a shared embedding space through the CLIP text encoder $g_t(\cdot)$:
\begin{equation}
    \label{eq: textual descriptors}
    \mathbf{z}_{i,m} = g_t(\mathbf{t}_{i,m}) \in \mathbb{R}^D, \quad m=1,\dots,M,
\end{equation}
which yields a set of textual embeddings that capture different levels of semantic abstraction, forming a hierarchical textual space that grounds the model for subsequent alignment with visual features. 

To enable such alignment, the vision encoder $g_v(\cdot)$ processes an input image $\mathbf{x}$ through $B$ layers, producing intermediate representations as defined in Eq.~\ref{eq:inter-visual-embeddings}. At each layer $b$, we extract the \verb|[CLS|] token $\mathbf{x}_{\text{cls}}^b \in \mathbb{R}^D$, which serves as the summary of visual information at the corresponding hierarchy of representation, and match it with the textual embedding $\mathbf{z}_{i,m}$ associated with class $i$. In other words, the collection of hierarchical visual embeddings $\{\mathbf{x}_{\text{cls}}^b\}_{b=1}^B$ is jointly aligned with the set of hierarchical textual descriptors $\{\mathbf{z}_{i,m}\}_{m=1}^M$, establishing cross-modal alignment across multiple hierarchical representations. We therefore compute the cosine similarity $s_{i,m}^b$ between $\mathbf{x}_{\text{cls}}^b$ and $\mathbf{z}_{i,m}$, which provides a quantitative measure of their alignment in the shared hierarchy space:
\begin{equation}
s_{i,m}^b = \cos\!\big(\mathbf{x}_{\text{cls}}^b, \mathbf{z}_{i,m}\big).
\label{eq: cos}
\end{equation}
However, simply using all descriptors directly may introduce redundancy and noise. To mitigate this, we retain only the Top-$K$ descriptors and normalize their weights. The aligned descriptors are then aggregated to form the united hierarchical textual embeddings for class $i$ at layer $b$:
\begin{align}
    \label{eq: router}
    \begin{aligned}
        &\mathbf{h}_i^b = \sum_{k} \alpha_{i,k}^b \,\mathbf{z}_{i,k}, \quad \text{where} \\
        &\alpha_{i,k}^b=\text{Softmax}\big(\text{Top}_K(\{s_{i, k}^b\}_{k=1}^K)\big).
    \end{aligned}
\end{align}
In Eq.~\ref{eq: router}, $\mathbf{h}_i^b$ represents a compact textual representation corresponding to $\mathbf{x}_{\text{cls}}^b$. By grounding intermediate visual features in their most relevant textual descriptors, the model benefits from hierarchical supervision that extends beyond the final-layer alignment in Eq.~\ref{eq:clip-loss}. As illustrated in Fig.~\ref{fig:overview}, this design enforces cross-modal alignment of hierarchical visual-textual representations, thereby preserving semantic coherence across levels, from coarse-grained cues to fine-grained details.

\subsection{Adaptive Hierarchy Routing}
While such hierarchical alignment establishes meaningful correspondences between intermediate visual and textual representations, an open question remains: how should these hierarchical representations be integrated to form the final prediction? A naïve strategy would be to average the representations across different layers, but this overlooks the fact that the relative importance of hierarchical levels can vary across tasks and inputs, with certain representations being more decisive for discrimination than others. To overcome this limitation, we introduce an adaptive routing mechanism that dynamically regulates the contributions of different hierarchical levels. Crucially, this routing must remain flexible as new tasks arrive, while also preserving knowledge from previously learned tasks to mitigate catastrophic forgetting.

Specifically, to adaptively regulate the relative contributions of hierarchical representations, we employ a router implemented as a lightweight linear layer $\mathbf{W}_r \in \mathbb{R}^{B\times D}$. Conditioned on the final-layer visual representation $\mathbf{x}_{\text{cls}}^B$, the router dynamically generates a set of non-negative weights $\boldsymbol{\beta}_i = \{\beta_i^1,\dots,\beta_i^B\}$ for the $B$ unified hierarchical textual embeddings of class $i$, and the final representation is subsequently obtained as their convex combination:
\begin{equation}
    \label{eq: convex}
    \tilde{\mathbf{h}}_i = \sum_{b=1}^B \beta_i^b \, \mathbf{h}_i^b, \quad \text{where} \ \boldsymbol{\beta}_i = \text{Softmax}\big(\mathbf{W}_r \mathbf{x}_{\text{cls}}^B\big).
\end{equation}
Therefore, the model can adaptively exploit different hierarchies depending on the input, thereby enhancing discrimination. Nonetheless, continually updating the router may overwrite routing patterns established for previous tasks, leading to catastrophic forgetting. To mitigate this, we update the router under a projection constraint that retains knowledge-rich subspaces, thus stabilizing learning across tasks. Specifically, after each incremental task, the weight of $\mathbf{W}_r$ is decomposed:
\begin{equation}
    \mathbf{W}_r = \mathbf{U} \mathbf{\Sigma} \mathbf{V}^\top,
\end{equation}
where $\mathbf{U} \in \mathbb{R}^{B \times B}$ and $\mathbf{V} \in \mathbb{R}^{D \times D}$ contain the left and right singular vectors, and $\mathbf{\Sigma} = \mathrm{diag}(\sigma_1,\ldots,\sigma_B)$ with $\sigma_b \ge 0$. We select the smallest rank $B'$ that preserves at least a proportion $\delta$ of the cumulative energy and form the projection matrix onto the retained subspace:
\begin{align}
    \label{eq: projection}
    \begin{aligned}
        &\mathbf{P}_{\text{old}} = \mathbf{U}_{[:,1:B']} \mathbf{U}_{[:,1:B']}^\top \in \mathbb{R}^{B \times B},\quad \text{where} \\
        &B' = \min \Bigg\{ r \;\Bigg|\; \frac{\sum_{b=1}^{r} \sigma_b^2}{\sum_{b=1}^{B} \sigma_b^2} \ge \delta \Bigg\}, \; \delta \in (0,1].
    \end{aligned}
\end{align}
where $\mathbf{P}_{\text{old}}$ acts as an orthogonal projector that encodes the routing-pattern subspace accumulated from previous tasks. During parameter updates, let \(\mathbf{W}_{\text{new}} \in \mathbb{R}^{B \times D}\) denote the router weight obtained by optimizing on the current task. We then decompose $\mathbf{W}_{\text{new}}$ into components inside and outside this preserved subspace and form the projected update using $\mathbf{P}_{\text{old}}$:
\begin{equation}
\label{eq: svd-update}
    \mathbf{W}_{\text{proj}} \;=\; \rho \, \mathbf{P}_{\text{old}} \, \mathbf{W}_{\text{new}} \;+\; (1-\rho)\, (\mathbf{I} - \mathbf{P}_{\text{old}})\, \mathbf{W}_{\text{new}},
\end{equation}
where $\mathbf{I} \in \mathbb{R}^{B \times B}$ is the identity matrix and $\rho \in [0,1]$ balances stability and plasticity. In Eq.~\ref{eq: svd-update}, the first term preserves and reinforces directions already captured by previous tasks, as it retains the component of $\mathbf{W}_{\text{new}}$ lying in the preserved subspace, whereas the second term instead extracts the component in the orthogonal complement, allowing the router to capture novel information and adapt to new tasks without disrupting prior knowledge. We then initialize the next task with $\mathbf{W}_{\text{proj}}$, which stabilizes routing dynamics across tasks and enables the model to flexibly integrate new knowledge while maintaining long-term consistency in continual learning.

\subsection{Cross-Modal Matching}
While the router integrates hierarchical textual representations, it remains essential to establish a joint space where visual and textual features can be directly compared and contrasted. To this end, we introduce an adapter $\mathbf{W}_a \in \mathbb{R}^{D\times D}$ that projects $\mathbf{x}_{\text{cls}}^B$ into the joint space, yielding $\mathbf{q}_i = \mathbf{W}_a \mathbf{x}_{\text{cls}}^B$ for cross-modal alignment. To enhance textual supervision, we mix the aggregated textual embedding $\tilde{\mathbf{h}}_i$ with the templated embedding $\mathbf{e}_i=g_t(\mathbf{t}_i)$ through a convex interpolation controlled by $\lambda \in [0,1]$. Building on this, we reformulate the prediction of class $i$ in Eq.~\ref{eq:clip-loss} using the adapted visual embedding and the mixed textual embedding, which yields the following contrastive loss:
\begin{align}
\label{eq:total-loss}
    \begin{aligned}
        &\mathcal{L}=\ell(f_{y_i}(\mathbf{x}), y_i),\quad\text{where} \\
        &f_{y_i}(\mathbf{x}) 
        = \frac{\exp\!\big(\cos(\mathbf{q}_i, (\lambda \tilde{\mathbf{h}}_i +(1-\lambda)\mathbf{e}_i)/\tau\big))}
        {\sum_{j=1}^{|\mathcal{Y}_t|} \exp\!\big(\cos(\mathbf{q}_i, (\lambda \tilde{\mathbf{h}}_j +(1-\lambda)\mathbf{e}_j)/\tau\big))}.
    \end{aligned}
\end{align}
To further mitigate forgetting, we adopt feature-level generative replay: rather than reconstructing raw images, we model the distribution of $\mathbf{x}_{\text{cls}}^B$ for each class $i$ as a Gaussian $\mathcal{N}(\boldsymbol{\mu}_i, \boldsymbol{\Sigma}_i)$ estimated from the empirical mean and covariance~\cite{DBLP:conf/cvpr/ZhuZWYL21}. For intermediate layers, only the diagonal covariance is retained to reduce memory cost. During incremental training, pseudo-features sampled from these Gaussians act as compact surrogates for past classes to alleviate forgetting.

\begin{table*}[t]
\small
\centering
\caption{{
Average and last performance comparison of different methods.  
        The best performance is shown in {\bf bold}.  
         All methods are initialized with the same pre-trained CLIP for a fair comparison.
}}
\renewcommand{\arraystretch}{1.1}
\setlength{\extrarowheight}{1pt}
\scriptsize{
\resizebox{\linewidth}{!}{
\begin{tabular}{l|cccc|cccc|cccc}
\hline
\rowcolor{gray!20}
 & 
\multicolumn{4}{c|}{\bm{{Aircraft}}} & 
\multicolumn{4}{c|}{\bm{{Cars}}} & 
\multicolumn{4}{c}{\bm{{CIFAR}}}
\\
\cline{2-13} 
\rowcolor{gray!20} &
\multicolumn{2}{c}{\bm{{B0\ Inc10}}} & 
\multicolumn{2}{c|}{\bm{{B50\ Inc10}}} & 
\multicolumn{2}{c}{\bm{{B0\ Inc10}}} & 
\multicolumn{2}{c|}{\bm{{B50\ Inc10}}} & 
\multicolumn{2}{c}{\bm{{B0\ Inc10}}} & 
\multicolumn{2}{c}{\bm{{B50\ Inc10}}} \\
\cline{2-13} 
\rowcolor{gray!20}
\multirow{-3}{*}{\bm{{Methods}}}  
& $\Bar{A}$ & $\mathcal{A}_B$ & $\Bar{A}$ & $\mathcal{A}_B$ & $\Bar{A}$ & $\mathcal{A}_B$ & $\Bar{A}$ & $\mathcal{A}_B$ & $\Bar{A}$ & $\mathcal{A}_B$ & $\Bar{A}$ & $\mathcal{A}_B$
\\
\hline
% Finetune & 3.16 & 0.96 & 1.72 & 1.05 & 3.14 & 1.10 & 1.54 & 1.13 & 7.84 & 4.44 & 5.30 & 2.46 \\
SimpleCIL~\cite{zhou2023revisiting} & 59.24 & 48.09 & 53.05 & 48.09 & 89.05 & 82.05 & 86.93 & 83.96 & 84.15 & 76.63 & 80.20 & 76.63 \\
\rowcolor{gray!10} CoOp~\cite{zhou2022learning} & 14.54 & 7.14 & 13.05 & 7.77 & 36.46 & 21.65 & 37.40 & 20.87 & 47.00 & 24.24 & 41.23 & 24.12 \\
ZS-CLIP~\cite{radford2021learning} & 26.66 & 17.22 & 21.70 & 17.22 & 82.60 & 76.37 & 78.32 & 76.37 & 81.81 & 71.38 & 76.49 & 71.38 \\
\rowcolor{gray!10} L2P~\cite{wang2022learning} & 47.19 & 28.29 & 44.07 & 32.13 & 76.63 & 61.82 & 76.37 & 65.64 & 82.74 & 73.03 & 81.14 & 73.61 \\
DualPrompt~\cite{DBLP:conf/eccv/0002ZESZLRSPDP22} & 44.30 & 25.83 & 46.07 & 33.57 & 76.26 & 62.94 & 76.88 & 67.55 & 81.63 & 72.44 & 80.12 & 72.57 \\
\rowcolor{gray!10} CODA-Prompt~\cite{DBLP:conf/cvpr/SmithKGCKAPFK23} & 45.98 & 27.69 & 45.14 & 32.28 & 80.21 & 66.47 & 75.06 & 64.19 & 82.43 & 73.43 & 78.69 & 71.58 \\
RAPF~\cite{huang2024class} & 50.38 & 23.61 & 40.47 & 25.44 & 82.79 & 71.22 & 77.21 & 69.97 & 86.14 & 78.04 & 82.17 & 77.93 \\
\rowcolor{gray!10} MG-CLIP~\cite{Huang_2025_ICCV} & 48.33 & 32.34 & 26.28 & 13.02 & 88.21 & 79.73 & 84.58 & 79.62 & 88.74 & 81.78 & 85.62 & 81.26 \\
PROOF~\cite{zhou2023learning} & 63.81 & 56.14 & 59.47 & 57.10 & 90.74 & 86.51 & 88.00 & 85.58 & 86.77 & 79.11 & 83.32 & 79.73 \\
\hdashline
\rowcolor{LightPurple}\momo & \bf66.70 & \bf58.84 & \bf60.41 & \bf58.18 & \bf92.49 &\bf 89.13 & \bf89.68 & \bf89.00 & \bf89.02 & \bf83.21 & \bf85.91 & \bf82.98 \\
\hline
\end{tabular}}}
\scriptsize{
\resizebox{\linewidth}{!}{
\begin{tabular}{l|cccc|cccc|cccc}
\rowcolor{gray!20}
 & 
\multicolumn{4}{c|}{\bm{{Food}}} & 
\multicolumn{4}{c|}{\bm{{UCF}}} & 
\multicolumn{4}{c}{\bm{{CUB}}}
\\
\cline{2-13} 
\rowcolor{gray!20} &
\multicolumn{2}{c}{\bm{{B0\ In10}}} & 
\multicolumn{2}{c|}{\bm{{B50\ Inc10}}} & 
\multicolumn{2}{c}{\bm{{B0\ In10}}} & 
\multicolumn{2}{c|}{\bm{{B50\ Inc10}}} & 
\multicolumn{2}{c}{\bm{{B0\ Inc20}}} & 
\multicolumn{2}{c}{\bm{{B100\ Inc20}}} \\
\cline{2-13} 
\rowcolor{gray!20}
\multirow{-3}{*}{\bm{{Methods}}}  
& $\Bar{A}$ & $\mathcal{A}_B$ & $\Bar{A}$ & $\mathcal{A}_B$ & $\Bar{A}$ & $\mathcal{A}_B$ & $\Bar{A}$ & $\mathcal{A}_B$ & $\Bar{A}$ & $\mathcal{A}_B$ & $\Bar{A}$ & $\mathcal{A}_B$
\\
\hline
% Finetune & 3.49 & 1.71 & 2.14 & 1.52 & 4.51 & 1.59 & 1.21 & 0.80 & 2.06 & 0.64 & 0.56 & 0.47 \\
SimpleCIL~\cite{zhou2023revisiting} & 87.89 & 81.65 & 84.73 & 81.65 & 90.44 & 85.68 & 88.12 & 85.68 & 82.80 & 76.21 & 78.01 & 74.72\\
\rowcolor{gray!10} CoOp~\cite{zhou2022learning} & 36.01 & 14.18 & 33.13 & 18.67 & 47.85 & 33.46 & 42.02 & 24.74 & 27.61 & 8.57 & 24.03 & 10.14 \\
ZS-CLIP~\cite{radford2021learning} & 87.86 & 81.92 & 84.75 & 81.92 & 75.50 & 67.64 & 71.44 & 67.64 & 74.38 & 63.06 & 67.96 & 63.06  \\
\rowcolor{gray!10} L2P~\cite{wang2022learning} & 85.66 & 77.33 & 80.42 & 73.13 & 86.34 & 76.43 & 83.95 & 76.62 & 70.87 & 57.93 & 75.64 & 66.12 \\
DualPrompt~\cite{DBLP:conf/eccv/0002ZESZLRSPDP22} & 84.92 & 77.29 & 80.00 & 72.75 & 85.21 & 75.82 & 84.31 & 76.35 & 69.89 & 57.46 & 74.40 & 64.84 \\
\rowcolor{gray!10} CODA-Prompt~\cite{DBLP:conf/cvpr/SmithKGCKAPFK23} & 86.18 & 78.78 & 80.98 & 74.13 & 87.76 & 80.14 & 83.04 & 75.03 & 73.12 & 62.98 & 73.95 & 62.21\\
RAPF~\cite{huang2024class} & 88.57 & 81.15 & 85.53 & 81.17 & 92.28 & 80.33 & 90.31 & 81.55 & 79.09 & 62.77 & 72.82 & 62.93\\
\rowcolor{gray!10} MG-CLIP~\cite{Huang_2025_ICCV} & 88.59 & 82.35 & 28.86 & 12.51 & 87.74 & 80.83 & 75.45 & 59.42 & 74.20 & 64.25 & 53.47 & 34.78 \\
PROOF~\cite{zhou2023learning} & 90.04 & 84.73 & 87.52 & 84.74 & 94.58 & 91.10 & 93.58 & 90.91 & 82.31 & 76.64 & 79.20 & 76.37 \\
\hdashline
\rowcolor{LightPurple}\momo & \bf90.93 & \bf86.18 & \bf88.53 & \bf86.26 & \bf97.69 & \bf95.72 & \bf96.63 & \bf95.83 & \bf84.71 & \bf78.37 & \bf80.52 & \bf77.95 \\
\hline
\end{tabular}}}
\scriptsize{
\resizebox{\linewidth}{!}{
\begin{tabular}{l|cccc|cccc|cccc}
\rowcolor{gray!20}
 & 
\multicolumn{4}{c|}{\bm{{ImageNet-R}}} & 
\multicolumn{4}{c|}{\bm{{ObjectNet}}} & 
\multicolumn{4}{c}{\bm{{SUN}}}
\\
\cline{2-13} 
\rowcolor{gray!20} &
\multicolumn{2}{c}{\bm{{B0\ In20}}} & 
\multicolumn{2}{c|}{\bm{{B100\ Inc20}}} & 
\multicolumn{2}{c}{\bm{{B0\ Inc20}}} & 
\multicolumn{2}{c|}{\bm{{B100\ Inc20}}} & 
\multicolumn{2}{c}{\bm{{B0\ Inc30}}} & 
\multicolumn{2}{c}{\bm{{B150\ Inc30}}} \\
\cline{2-13} 
\rowcolor{gray!20}
\multirow{-3}{*}{\bm{{Methods}}}  
& $\Bar{A}$ & $\mathcal{A}_B$ & $\Bar{A}$ & $\mathcal{A}_B$ & $\Bar{A}$ & $\mathcal{A}_B$ & $\Bar{A}$ & $\mathcal{A}_B$ & $\Bar{A}$ & $\mathcal{A}_B$ & $\Bar{A}$ & $\mathcal{A}_B$
\\
\hline
% Finetune & 1.37 & 0.43 & 1.01 & 0.88 & 1.34 & 0.47 & 0.69 & 0.54 & 4.51 & 1.59 & 0.78 & 0.72 \\
SimpleCIL~\cite{zhou2023revisiting} & 81.06 & 74.48 & 76.84 & 74.48 & 52.06 & 40.13 & 45.11 & 40.13 & 82.13 & 75.58 & 78.62 & 75.58 \\
\rowcolor{gray!10} CoOp~\cite{zhou2022learning} & 60.73 & 37.52 & 54.20 & 39.77 & 21.24 & 6.29 & 16.21 & 6.82 & 45.93 & 23.11 & 39.33 & 24.89 \\
ZS-CLIP~\cite{radford2021learning} & 83.37 & 77.17 & 79.57 & 77.17 & 38.43 & 26.43 & 31.12 & 26.43 & 79.42 & 72.11 & 74.95 & 72.11 \\
\rowcolor{gray!10} L2P~\cite{wang2022learning} & 75.97 & 66.52 & 72.82 & 66.77 & 51.40 & 39.39 & 48.91 & 42.83 & 82.82 & 74.54 & 79.57 & 73.10 \\
DualPrompt~\cite{DBLP:conf/eccv/0002ZESZLRSPDP22} & 76.21 & 66.65 & 73.22 & 67.58 & 52.62 & 40.72 & 49.08 & 42.92 & 82.46 & 74.40 & 79.37 & 73.02 \\
\rowcolor{gray!10} CODA-Prompt~\cite{DBLP:conf/cvpr/SmithKGCKAPFK23} & 77.69 & 68.95 & 73.71 & 68.05 & 46.49 & 34.13 & 40.57 & 34.13 & 83.34 & 75.71 & 80.38 & 74.17\\
RAPF~\cite{huang2024class} & 83.56 & 76.63 & 79.61 & 75.92 & 53.78 & 34.97 & 45.37 & 35.74 & 85.23 & 78.21 & 81.91 & 78.62 \\
\rowcolor{gray!10} MG-CLIP~\cite{Huang_2025_ICCV} & 83.18 & 77.20 & 50.47 & 35.37 & 51.41 & 40.90 & 25.00 & 12.39 & 53.71 & 41.62 & 26.58 & 12.64 \\
PROOF~\cite{zhou2023learning} & 83.84 & 78.40 & 81.20 & 78.92 & 56.07 & 43.69 & 48.90 & 43.62 & 83.89 & 77.25 & 80.15 & 76.54 \\
\hdashline
\rowcolor{LightPurple}\momo & \bf84.98 & \bf80.73 & \bf82.04 & \bf80.70 & \bf56.79 & \bf44.21 & \bf49.78 & \bf44.34 & \bf86.76 & \bf80.76 & \bf83.51 & \bf80.85\\
\hline
\end{tabular}}}
\vspace{-2mm}
\label{tab:benchmark}
\end{table*}

\begin{figure*}
	\centering
 \vspace{-4mm}
	\begin{subfigure}{0.32\linewidth}
		\includegraphics[width=1\linewidth]{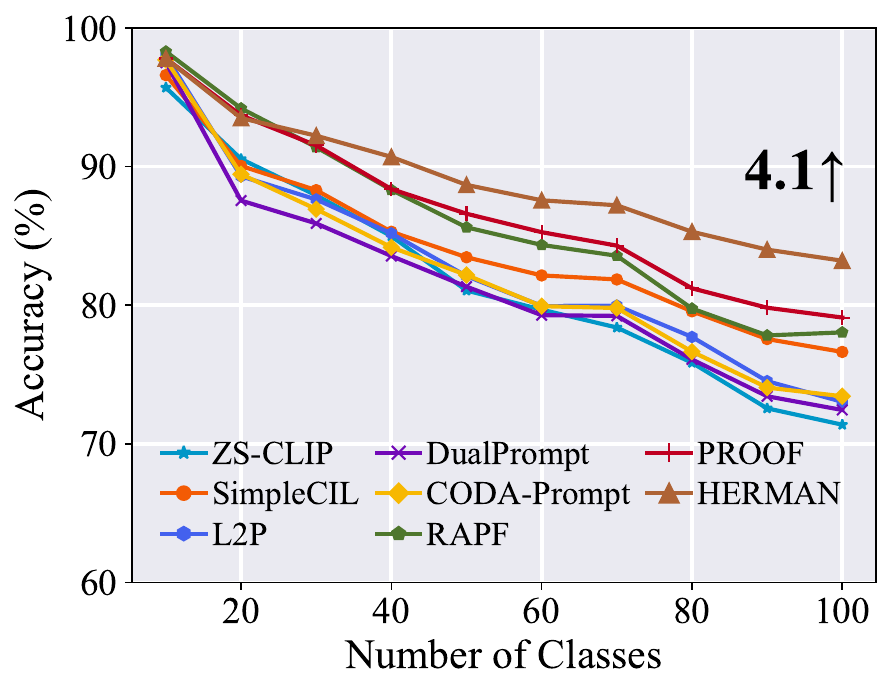}
		\caption{CIFAR Base0 Inc10}
		\label{fig:benchmark-cifar}
	\end{subfigure}
	\hfill
     \begin{subfigure}{0.32\linewidth}
		\includegraphics[width=1\linewidth]{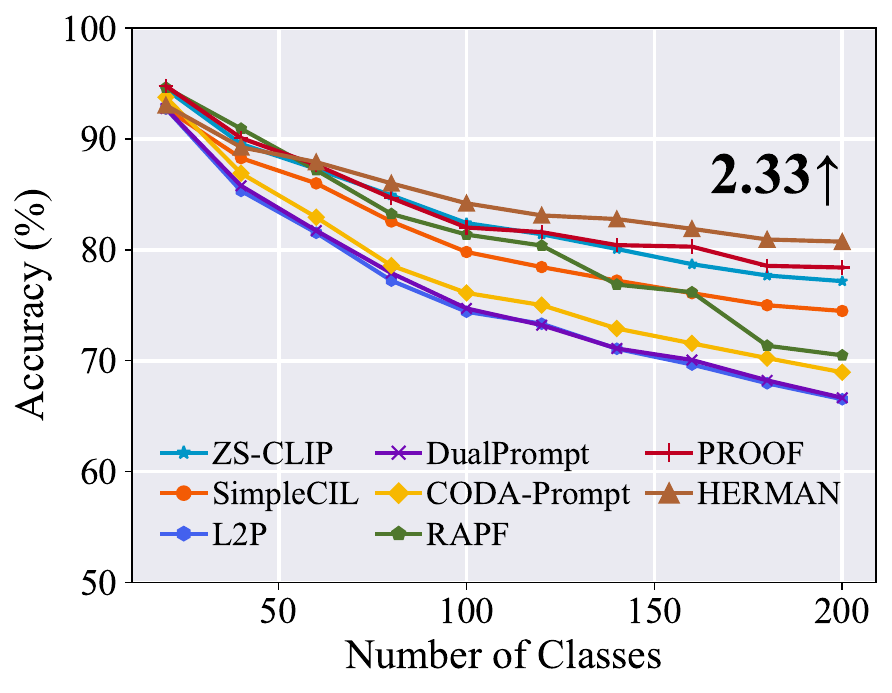}
		\caption{ImageNet-R Base0 Inc10}
		\label{fig:benchmark-imagenetr}
	\end{subfigure}
	\hfill
	\begin{subfigure}{0.32\linewidth}
		\includegraphics[width=1\linewidth]{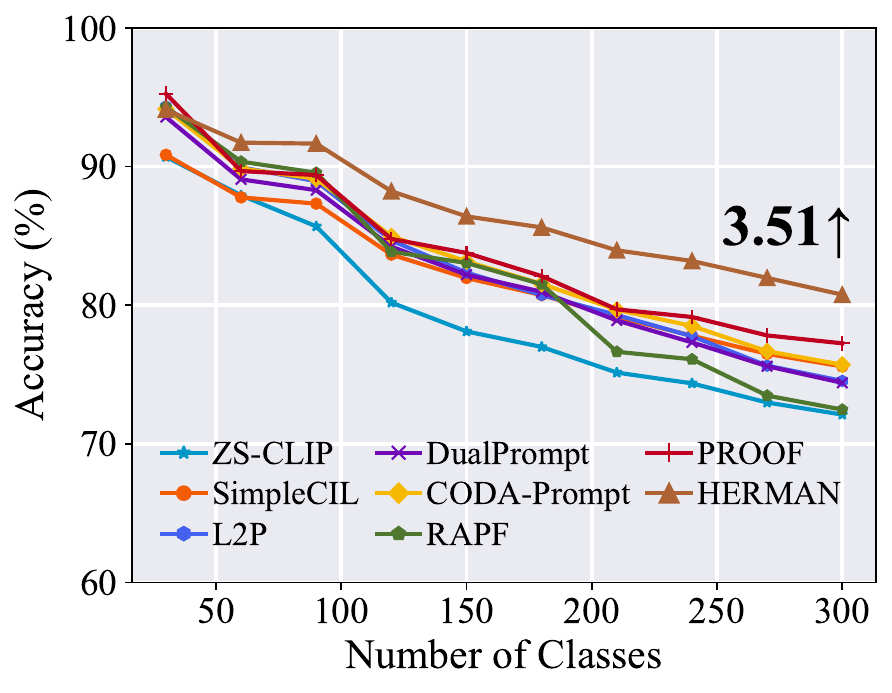}
		\caption{SUN Base0 Inc10}
		\label{fig:benchmark-sun}
	\end{subfigure}
	\caption{Incremental performance of different methods. We report the performance gap after the last incremental stage of \momo\ and the runner-up method at the end of the line. All methods utilize the same CLIP pre-trained weights.}
	\label{fig:benchmark}
 \vspace{-5mm}
\end{figure*}

\textbf{Summary of \momo.}
We enhance CLIP-based CIL by introducing LLM-generated hierarchical textual descriptors, which are encoded into embeddings with Eq.~\ref{eq: textual descriptors}, aligned with intermediate visual features with Eq.~\ref{eq: cos}, and further aggregated into unified hierarchical representations with Eq.~\ref{eq: router}. A router convexly combines these embeddings with Eq.~\ref{eq: convex} and is updated with a projection constraint to effectively mitigate forgetting with Eq.~\ref{eq: svd-update}. During training, the model is jointly optimized with Eq.~\ref{eq:total-loss}. During inference, final predictions are obtained by matching visual embeddings against the mixed textual embeddings of all previously seen classes using the same formulation.

\begin{table*}[t]
\small
\centering
\caption{Results when all methods use the same textual descriptors generated by different LLMs.}
\renewcommand{\arraystretch}{1.1}
\setlength{\extrarowheight}{1pt}
\scriptsize{
\resizebox{\linewidth}{!}{
\begin{tabular}{l|c|cc|cc|cc|cc}
\hline
\rowcolor{gray!20}
            && 
			\multicolumn{2}{c | }{\bf Aircraft B0 Inc10} & \multicolumn{2}{c | }{\bf CIFAR B0 Inc10} & \multicolumn{2}{c | }{\bf UCF B0 Inc10} & \multicolumn{2}{c }{\bf CUB B0 Inc20}  \\
\cline{3-10} 
	\rowcolor{gray!20}
\multirow{-2}{*}{\bm{{Methods}}} & \multicolumn{1}{c | }{\multirow{-2}{*}{\bf Textual Descriptors}}  & {$\bar{\mathcal{A}}$} & ${\mathcal{A}_B}$  
			& {$\bar{\mathcal{A}}$} & ${\mathcal{A}_B}$
            & {$\bar{\mathcal{A}}$} & ${\mathcal{A}_B}$
            & {$\bar{\mathcal{A}}$} & ${\mathcal{A}_B}$ \\
\hline
            ZS-CLIP~\cite{radford2021learning} & GPT-5 Generated & 31.33 & 22.69 & 84.07 & 75.50 & 80.58 & 71.55 & 79.97 & 68.43 \\
			\rowcolor{gray!10} RAPF~\cite{huang2024class} & GPT-5 Generated & 55.71 & 32.96 & 88.35 & 81.31  & 94.52 & 86.69 & 82.48 & 65.95\\
			\rowcolor{LightPurple}	\momo  & GPT-5 Generated & \bf 66.70 & \bf58.84  & \bf89.02 & \bf83.21  & \bf97.69 & \bf95.72 & \bf84.71 & \bf78.37\\
			\midrule
			ZS-CLIP~\cite{radford2021learning} & QWEN-Plus Generated & 31.25 & 22.54 & 84.24 & 75.68 & 80.63 & 71.55 & 79.89 & 68.31 \\
			\rowcolor{gray!10} RAPF~\cite{huang2024class} & QWEN-Plus Generated & 55.75 & 32.92 & 88.41 & 81.32  & 94.49 & 86.60 & 83.13 & 65.02\\
			\rowcolor{LightPurple}	\momo  & QWEN-Plus Generated & \bf66.60& \bf58.30  & \bf89.02 & \bf83.17  & \bf94.34 & \bf90.60 & \bf84.87 & \bf79.86\\
\hline
\end{tabular}}}
\label{tab:benchmark-prompt}
\end{table*}

\section{Experiments}
\label{experiments}
In this section, we evaluate \momo\ on nine benchmark datasets and compare it with state-of-the-art approaches. In addition, we conduct ablation studies and parameter sensitivity analysis, and provide visualizations to further validate the robustness and interpretability of our framework.

\subsection{Implementation Details}
\noindent\textbf{Dataset.} We follow~\cite{zhou2023learning,zhou2022learning,wang2022learning} to evaluate the performance on nine benchmark datasets that have domain gap to CLIP's pre-training dataset, \textit{i.e.}, {CIFAR100}~\cite{Krizhevsky2009LearningML}, {CUB200}~\cite{WahCUB2002011}, {ObjectNet}~\cite{barbu2019objectnet}, {ImageNet-R}~\cite{hendrycks2021many}, {FGVCAircraft}~\cite{maji2013fine}, {StanfordCars}~\cite{krause20133d}, {Food101}~\cite{bossard2014food}, {SUN397}~\cite{xiao2010sun} and {UCF101}~\cite{soomro2012ucf101}.

\noindent\textbf{Dataset Split.} Following~\cite{rebuffi2017icarl,wang2022learning}, we use `B-$m$ Inc-$n$' to split the classes in CIL. $m$ indicates the number of classes in the first stage, and $n$ represents that of every following stage. The dataset split is adapted following \cite{zhou2023learning}. We follow~\cite{rebuffi2017icarl} to randomly shuffle the class order with random seed 1993 for all experiments.

\noindent\textbf{Comparison Methods.} We first compare to SOTA pre-trained model-based CIL algorithms, \textit{e.g.}, L2P~\cite{wang2022learning}, DualPrompt~\cite{DBLP:conf/eccv/0002ZESZLRSPDP22}, CODA-Prompt~\cite{DBLP:conf/cvpr/SmithKGCKAPFK23} and SimpleCIL~\cite{zhou2023revisiting}. Besides, we also compare to SOTA CLIP-based CIL algorithms, \textit{e.g.}, CoOp~\cite{zhou2022learning}, RAPF~\cite{huang2024class} and PROOF~\cite{zhou2023learning}. As a baseline, we include a variant that fine-tunes CLIP on incremental tasks, denoted as Finetune. All methods are deployed with the same CLIP as initialization.

\noindent\textbf{Training Details.} All experiments are conducted on NVIDIA RTX 4090 GPUs using PyTorch~\cite{paszke2019pytorch}. Following~\cite{huang2024class, zhou2023learning}, we adopt CLIP with a ViT-B/16 backbone pre-trained on LAION-400M~\cite{radford2021learning} as the foundation model for all compared methods to ensure a fair comparison. For vision-only baselines that cannot directly leverage textual prompts (\textit{e.g.}, L2P, DualPrompt, CODA-Prompt), we initialize their encoders using only the visual branch of CLIP. In \momo, we optimize the model using SGD with a batch size of $64$ for $20$ epochs, starting from an initial learning rate of $0.05$ and decaying it according to the prescribed schedule. We set $\text{Top-}k=5$ and $\lambda=0.5$ for embedding combination, and use $\rho=0.9$ and $\delta=0.9$ to control the router weight updates. GPT-5~\cite{OpenAI2025GPT5} is employed to generate textual descriptors that enrich the semantic space used for hierarchical representation matching.

\noindent\textbf{Evaluation Metrics.}
Following~\cite{rebuffi2017icarl, zhou2023learning}, we use $\mathcal{A}_t$ to represent the model's accuracy after the $t$-th task. Specifically, we adopt $\mathcal{A}_T$ (the performance after the last task) and $\bar{\mathcal{A}}=\frac{1}{T}\sum_{t=1}^{T}\mathcal{A}_t$ (average performance along incremental tasks) as measurements.

\subsection{Benchmark Comparison and Further Analysis}

\noindent\textbf{Benchmark Comparison.} We first compare \momo\ with state-of-the-art methods on benchmark datasets, as reported in Tab.~\ref{tab:benchmark} and Fig.~\ref{fig:benchmark}. \momo\ consistently outperforms existing approaches by $1\%$–$5\%$. The Finetune baseline performs the worst, reflecting its severe forgetting of fine-grained class features. Visual prompt-based methods (\textit{e.g.}, L2P, DualPrompt, CODA-Prompt) are limited by their inability to leverage textual semantics, while the textual prompt tuning method CoOp suffers from forgetting of learned prompts, leading to suboptimal results. Notably, \momo\ even surpasses the exemplar-based method PROOF, further demonstrating its effectiveness. This advantage stems from hierarchical representation matching, which aligns multi-level textual descriptors with visual features, enabling stronger cross-modal alignment and more stable knowledge retention.

\noindent\textbf{Robustness for Textual Descriptors.} We investigate how \momo\ performs using different LLM-generated textual descriptors. Specifically, we apply the same prompt for GPT-5~\cite{OpenAI2025GPT5} and QWEN-Plus~\cite{qwen3} to generate textual descriptors, and report the results in Tab.~\ref{tab:benchmark-prompt}. Remarkably, even with the same prompt, \momo\ consistently outperforms the other methods. This demonstrates that \momo\ not only benefits from hierarchical representations but also exhibits robustness in leveraging textual descriptors generated by different LLMs.

\begin{figure*}
% \vspace{-3mm}
	\centering
	\begin{subfigure}{0.38\linewidth}
		\includegraphics[width=1\linewidth]{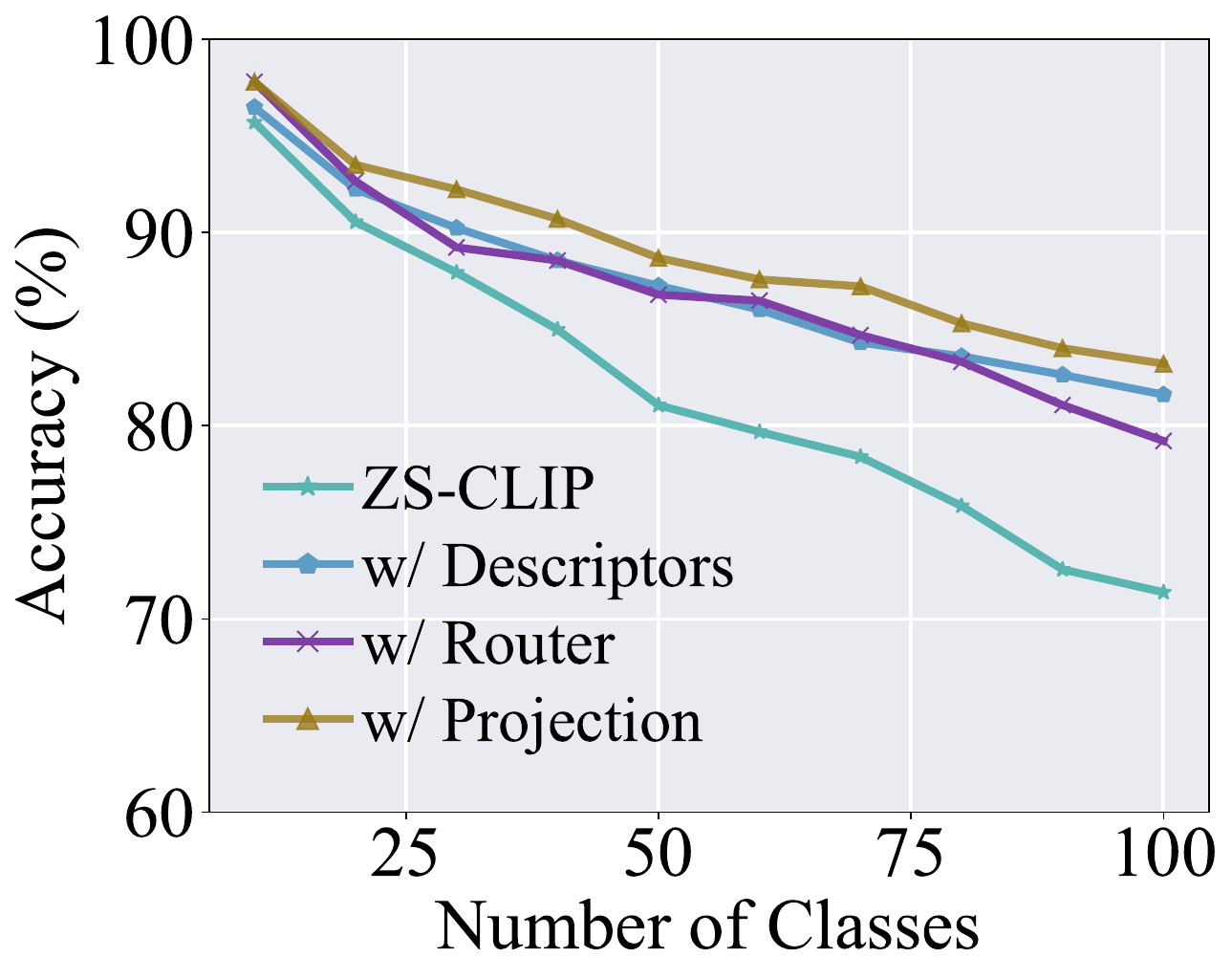}
		\caption{Ablation study}
      \label{ablation}
	\end{subfigure}
	\hfill
	\begin{subfigure}{0.30\linewidth}
		\includegraphics[width=1\linewidth]{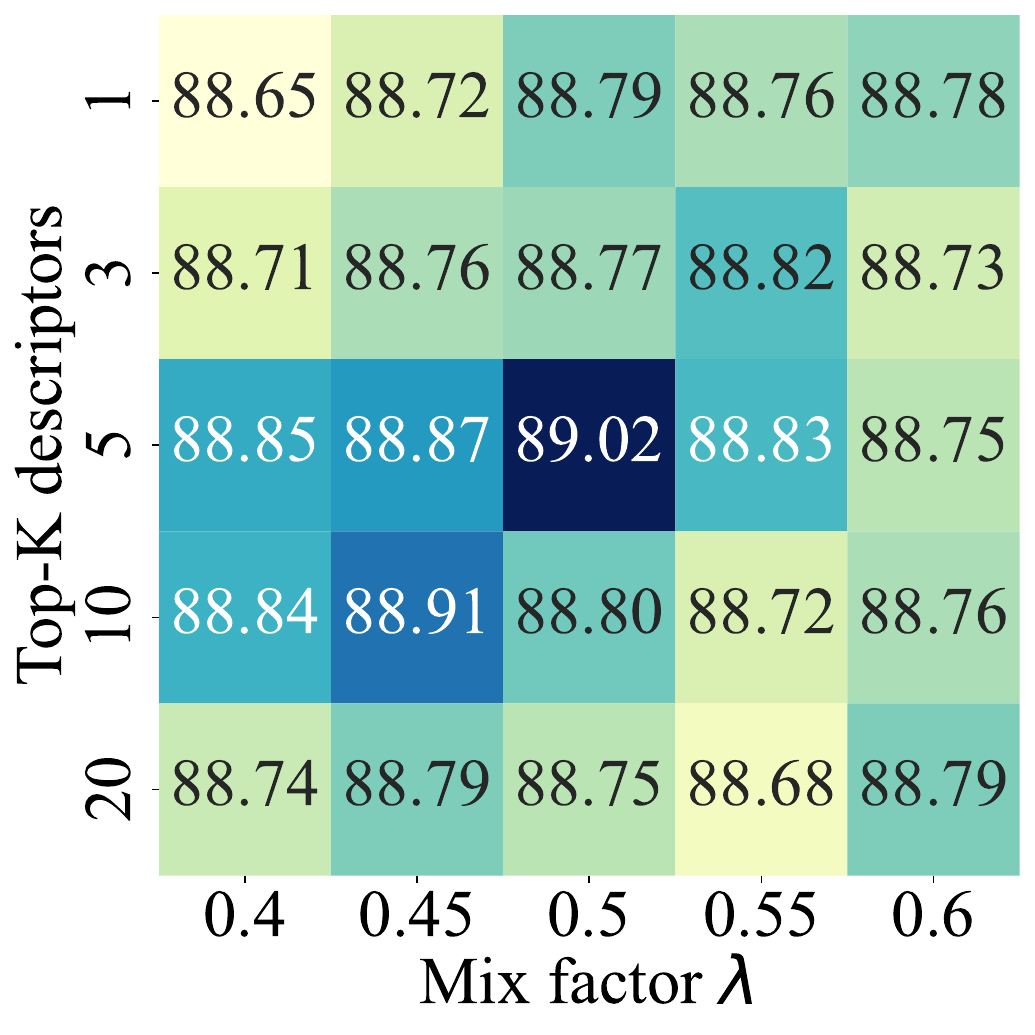}
		\caption{Parameter sensitivity for $K$ and $\lambda$}
  \label{sensitivity}
	\end{subfigure}
	\hfill
	\begin{subfigure}{0.30\linewidth}
		\includegraphics[width=1\linewidth]{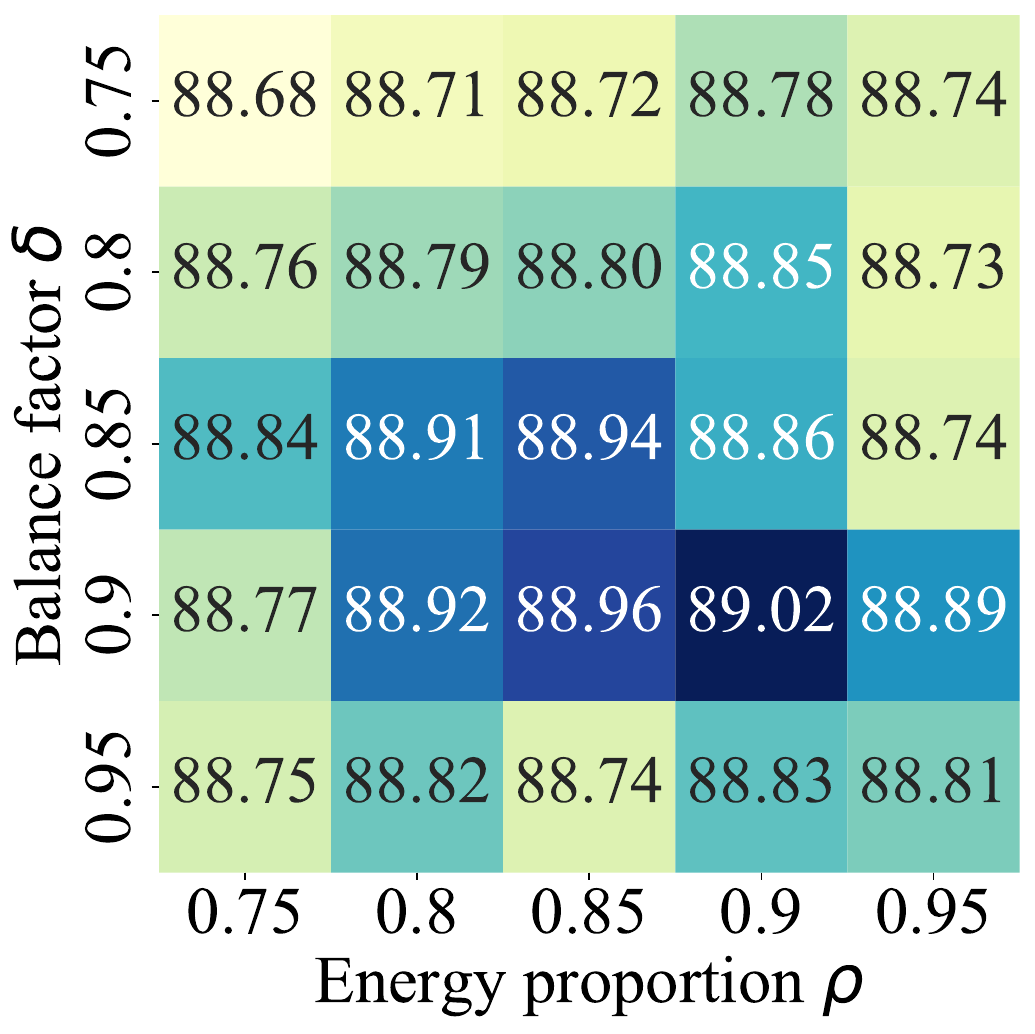}
		\caption{Parameter sensitivity for $\delta$ and $\rho$}
  \label{routersensitivity}
	\end{subfigure}
	\caption{Ablation study and parameter sensitivity analysis. The results demonstrate \momo's robustness to variations in key hyperparameters, with stable performance across a range of settings.}
	\label{fig:ablandsen}
 \vspace{-5mm}
\end{figure*}

\noindent\textbf{Ablation Study.} We conduct ablation experiments on CIFAR100 B0 Inc10 to evaluate the contribution of each component in \momo, with results shown in Fig.~\ref{ablation}. {\bf `ZS-CLIP'} serves as the baseline and performs the worst due to distributional shifts. {\bf `w/ Descriptors'} introduces LLM-generated textual descriptors (Eq.~\ref{eq: textual descriptors}) and significantly boosts performance, showing the benefit of richer features. {\bf `w/ Router'} adds a router without constraints (Eq.~\ref{eq: router}) but provides little or no improvement. Finally, {\bf `w/ Projection'}, the full \momo\ model with projection-constrained router updates (Eq.~\ref{eq: projection}), achieves the best results, confirming the effectiveness of each component.

\noindent\textbf{Parameter Robustness.} We evaluate robustness on CIFAR100 B0 Inc10 by varying two parameters: the number of top-$K$ textual descriptors and the mixing factor $\lambda$ between templated and unified embeddings. Specifically, we set $K \in \{1,3,5,10,20\}$ and $\lambda \in \{0.40, 0.45, 0.50, 0.55, 0.60\}$. The average performance $\bar{\mathcal{A}}$ is shown in Fig.~\ref{sensitivity}. Results indicate that too few descriptors fail to capture sufficient hierarchical information, while too many introduce noise. In contrast, $\lambda$ shows stable performance across values. Overall, \momo\ exhibits strong robustness to these hyperparameter choices, and we recommend $K=5$ and $\lambda=0.5$ as default settings.

\begin{figure*}[t]
% \vspace{-3mm}
	\centering
	\begin{subfigure}[t]{0.32\linewidth}
    \vspace{0pt}
		\includegraphics[width=1\linewidth]{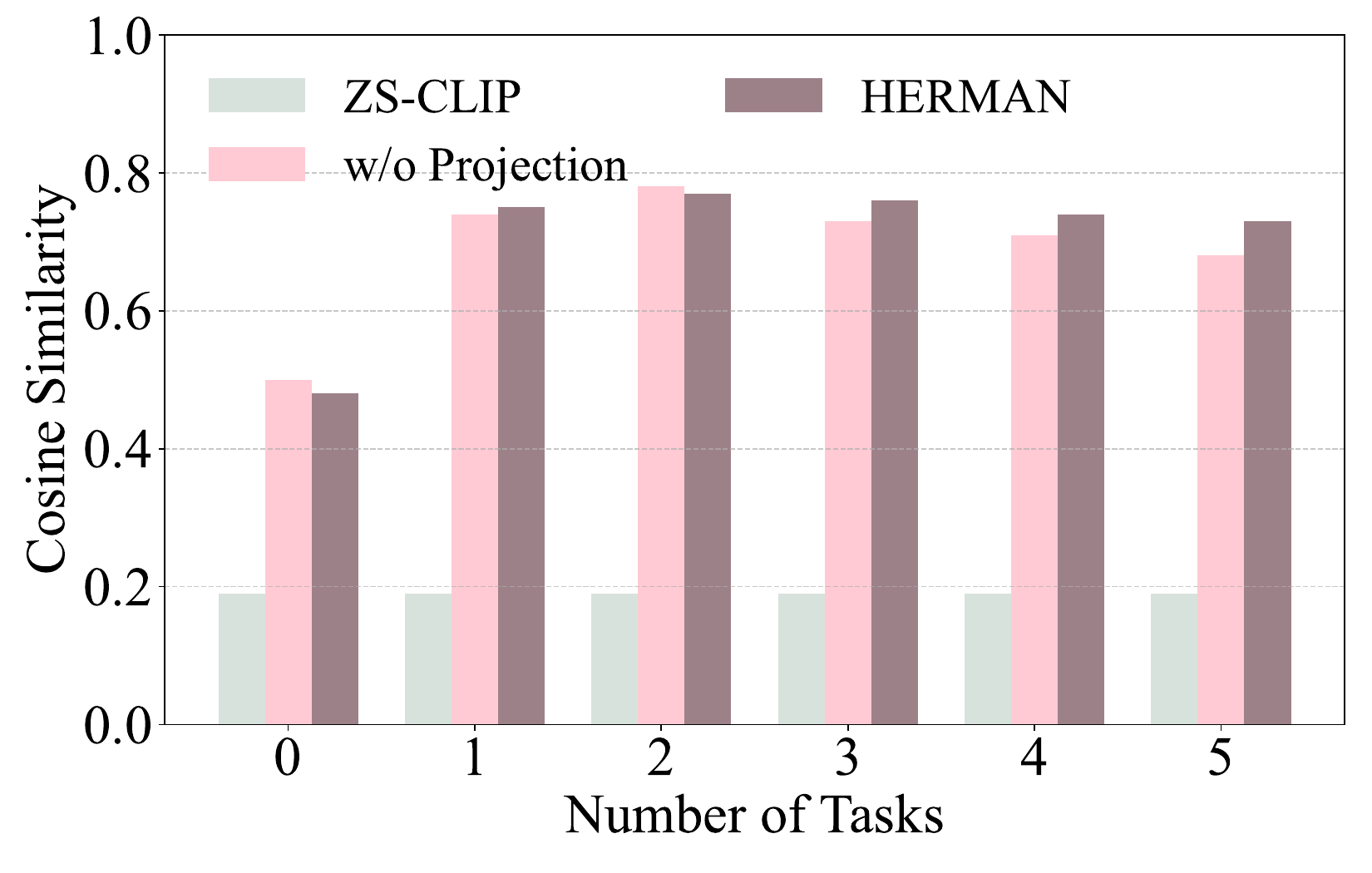}
		\caption{Cosine similarity analysis}
  \label{similarity}
	\end{subfigure}
 \hfill
	\begin{subfigure}[t]{0.32\linewidth}
    \vspace{0pt}
		\includegraphics[width=1\linewidth]{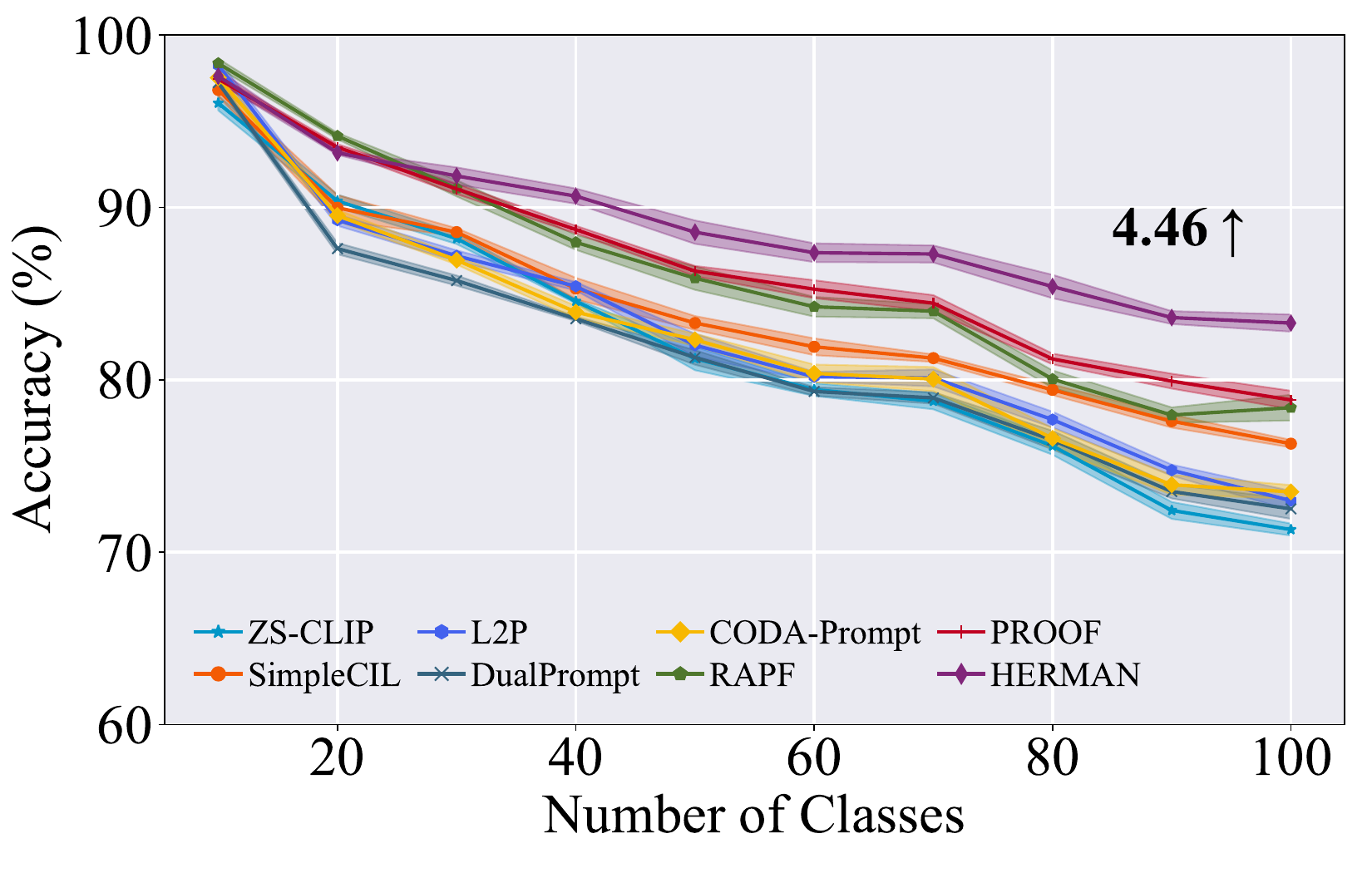}
		\caption{Evaluation across multiple random seeds}
  \label{Multiple}
	\end{subfigure}
 \hfill
	\begin{subfigure}[t]{0.32\linewidth}
    \vspace{0pt}
		\includegraphics[width=1\linewidth]{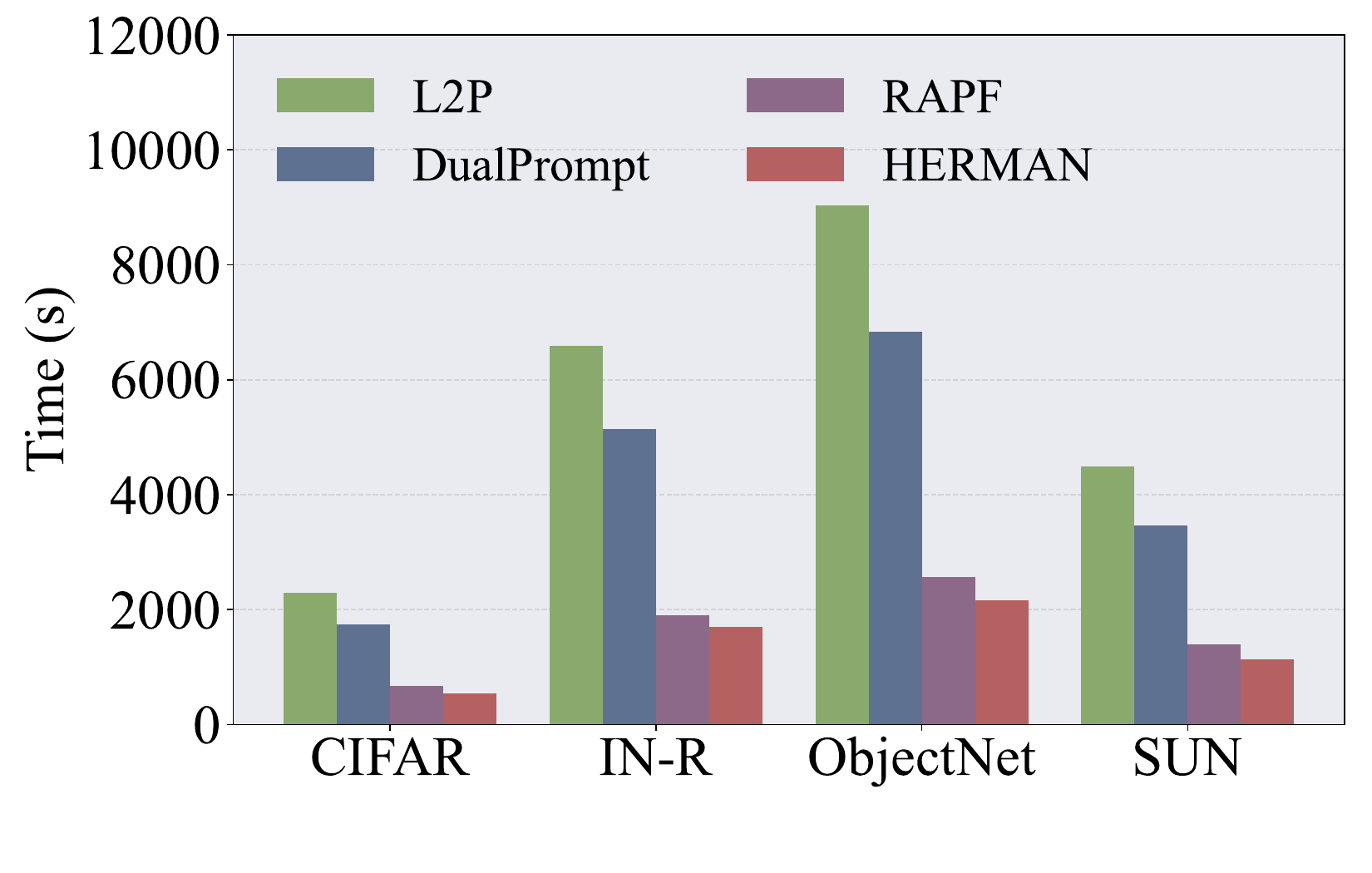}
		\caption{Running time comparison}
  \label{runtime}
	\end{subfigure}
	\caption{Evaluation of \momo\ across different aspects: (a) Cosine similarity analysis, (b) Evaluation across multiple random seeds, and (c) Running time comparison on four datasets.}
	\label{fig:sim_multiple_time}
 \vspace{-2mm}
\end{figure*}

\begin{figure*}
	\centering
	\begin{subfigure}{0.32\linewidth}
		\includegraphics[width=1\linewidth]{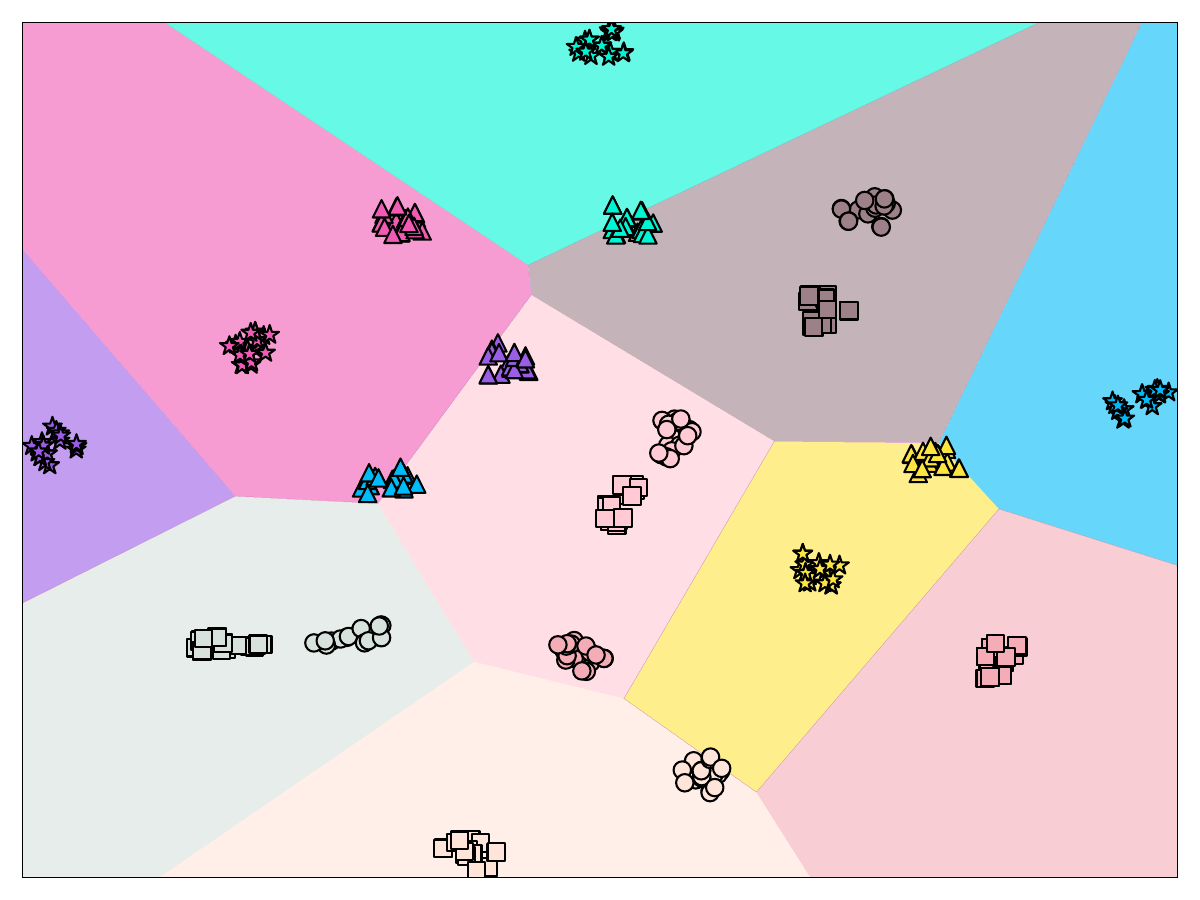}
		\caption{Matching with 4 layers}
	\end{subfigure}
	\hfill
	\begin{subfigure}{0.32\linewidth}
		\includegraphics[width=1\linewidth]{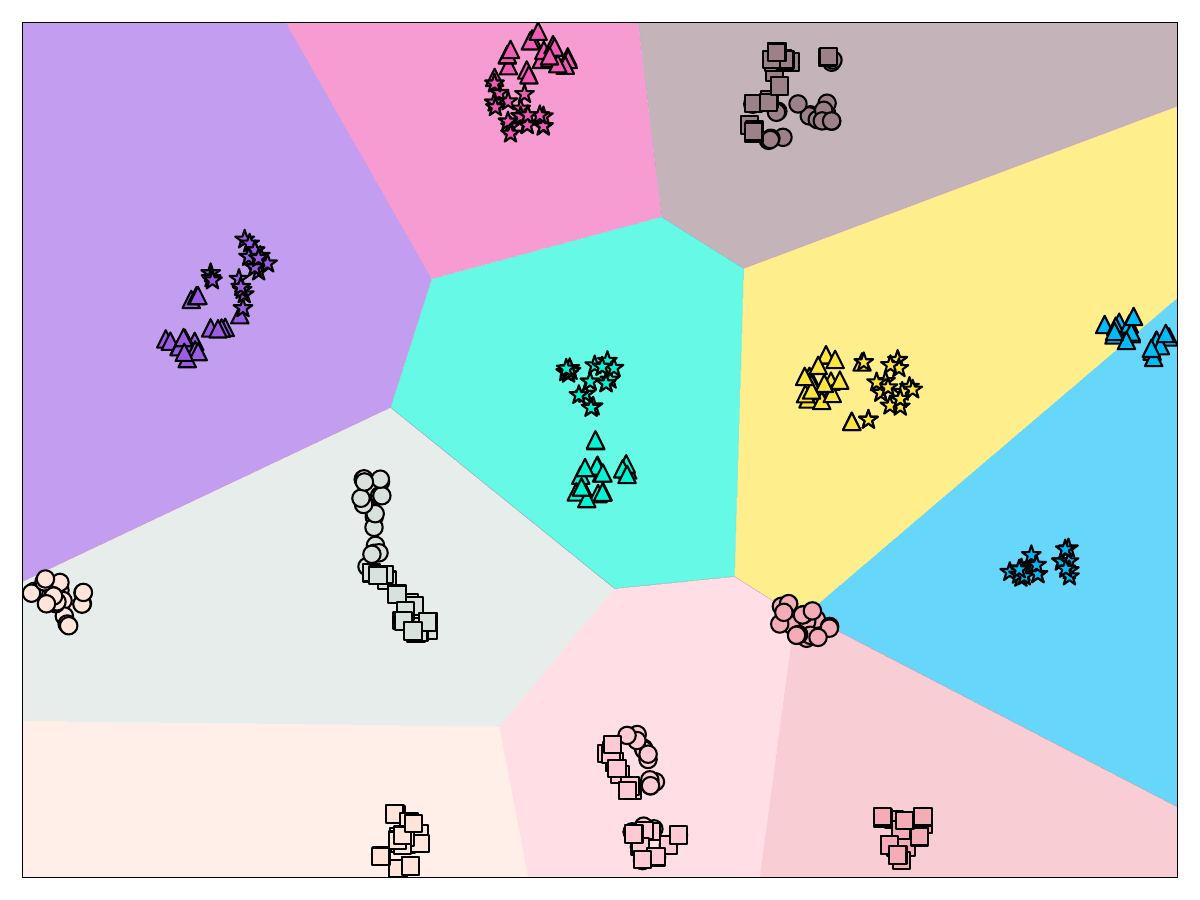}
		\caption{Matching with 8 layers}
	\end{subfigure}
	\hfill
	\begin{subfigure}{0.32\linewidth}
		\includegraphics[width=1\linewidth]{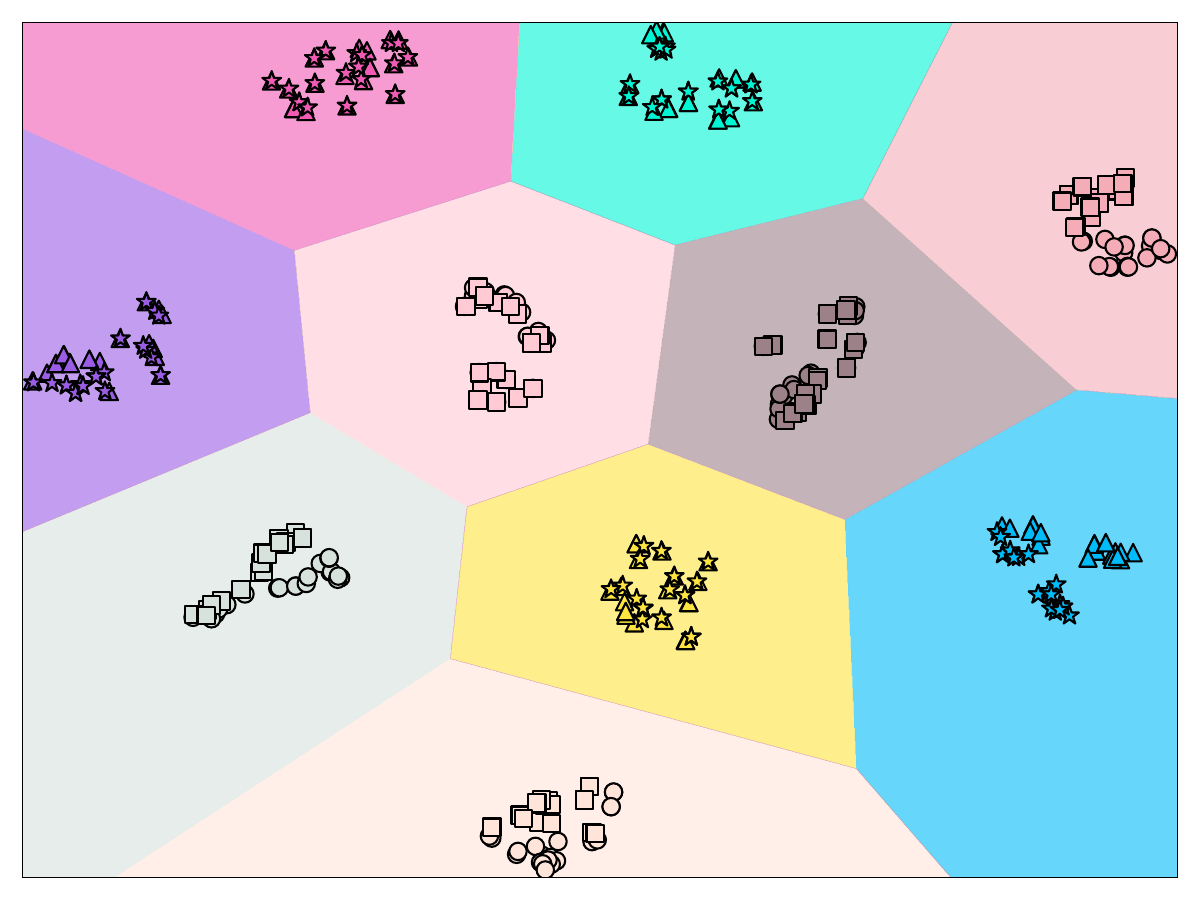}
		\caption{Matching with 12 layers}
	\end{subfigure}
	\caption{We visualize 5 classes of the first task in light colors. In this case, visual and textual embeddings are represented by circles and squares. For the second task, 5 classes are shown in vivid colors, with visual and textual embeddings represented by triangles and stars, respectively. All embeddings are visualized after the completion of all learning tasks.}
	\label{fig:tsne} 
 \vspace{-5mm}
\end{figure*}

We further evaluate the robustness of \momo\ with respect to the balance factor $\delta$ in the projection step and the mixing coefficient $\rho$, with results summarized in Fig.~\ref{routersensitivity}. As shown, varying $\delta$ across $\{0.75, 0.80, 0.85, 0.90, 0.95\}$ and $\rho$ across $\{0.75, 0.80, 0.85, 0.90, 0.95\}$ leads to only minor fluctuations in performance, all remaining within a narrow band around the reported average accuracy. This demonstrates that the model is not overly sensitive to either parameter: $\delta$ effectively controls the proportion of preserved subspace without destabilizing learning, while $\rho$ balances stability and plasticity in a consistent manner. Overall, \momo\ shows strong robustness under different choices of $\delta$ and $\rho$, with default settings $\delta=0.9$ and $\rho=0.9$ adopted throughout the main experiments.

\noindent\textbf{Update Router without Forgetting.} We conduct experiments on CIFAR100 B50 Inc10 to investigate the role of $\mathbf{P}_{\textbf{old}}$ in maintaining the routing pattern subspace learned from earlier tasks. The cosine similarity between the visual and textual embeddings of the first-task classes is averaged, as shown in Fig.~\ref{similarity}. {\bf `ZS-CLIP'} freezes the model without adaptation, yielding no similarity improvement. {\bf `w/o Projection'}, which updates the router without projecting $\mathbf{P}_{\textbf{old}}$, initially adapts to downstream tasks but quickly deteriorates due to the lack of constraints to retain prior routing patterns. In contrast, {\bf `HERMAN'} leverages the projection mechanism during router updates and sustains the highest cosine similarity as tasks are added incrementally, highlighting the effectiveness of the projection constraint in mitigating catastrophic forgetting.

\noindent\textbf{Evaluation Across Multiple Random Seeds.}
To further assess the robustness and stability of different approaches, we additionally perform multiple independent runs by varying the random seed used for class partitioning. Specifically, we conduct five separate trials with random seeds \{1991, 1992, 1993, 1994, 1995\} and report the mean accuracy as well as the standard deviation across runs. As shown in Fig.~\ref{Multiple}, \momo\ exhibits consistently superior robustness compared to prior methods, achieving both higher average accuracy and lower variance across different random seeds. These results indicate that \momo\ remains stable under varied class-split configurations and demonstrates strong reliability in practical continual learning scenarios.

\noindent\textbf{Running Time Consumption.}
We provide a comprehensive analysis of the running time across different methods evaluated on four datasets: CIFAR, IN-R, ObjectNet and SUN. All experiments are conducted on a single NVIDIA 4090 GPU to ensure a fair comparison. As shown in Fig.~\ref{runtime}, our method consistently requires less training time than the other competing methods while maintaining superior performance. Notably, our method outperforms L2P, DualPrompt, and RAPF by a significant margin on all datasets, with particularly large improvements observed on larger datasets such as ObjectNet and SUN. These results highlight the efficiency of our method, which demonstrates the best computational cost across all evaluated datasets. This confirms that our method not only achieves the best performance but also minimizes computational overhead, making it both efficient and effective for large-scale continual learning tasks.

\noindent\textbf{Effect of Hierarchical Representations Matching.}
We investigate the impact of hierarchical information on cross-modal alignment using CIFAR100 B0 Inc10 with t-SNE~\cite{maaten2008visualizing}, as shown in Fig.~\ref{fig:tsne}. Alignment is performed with textual descriptors extracted from 4 to 12 intermediate layers. As more layers are incorporated, embeddings of visual and textual modalities cluster more tightly and their gap progressively narrows, indicating stronger alignment. This trend demonstrates that richer hierarchical information consistently enhances cross-modal matching.

\noindent\textbf{Prompts for Generating Textual Descriptions.}
In this work, we leverage prompt-based learning to generate textual descriptions for different categories within a vision-language framework. We present a sample prompt as shown in Code~\ref{lst:python_example}. Prompts are designed to guide the model in producing task-specific descriptions that are both relevant and informative for each class. We focus on generating detailed, yet concise, visual descriptions that can effectively distinguish between different object categories. The generation process follows a template-driven approach, where the input prompt consists of predefined instructions and placeholders for category names. The model is then tasked with completing the description based on these instructions. These instructions guide the model to generate descriptions that are not only accurate but also aligned with human perception, providing a diverse set of visual features that can be used for downstream tasks such as object recognition and classification. The prompts are carefully designed to ensure that the generated descriptions are useful and concise. For example, in the case of an apple, the generated descriptions might include characteristics such as ``round or oval shape'', ``smooth surface'', and ``presence of a stem''. Each description is intended to capture a distinctive feature of the object that is easily observable in real-world images, making the generated descriptions both practical and effective for distinguishing between categories in a variety of computer vision tasks. By using prompt-based generation, we achieve a flexible and efficient way to produce textual descriptions without relying on large labeled datasets or manually curated feature sets.

\begin{figure*}[ht]
    \centering
\begin{lstlisting}[language=Python, caption=A Python function for generating visual descriptions to distinguish object categories in images with MLLMs., label={lst:python_example}, basicstyle=\ttfamily\small, frame=single, backgroundcolor=\color{lightgray}]
def generate_descriptions(category: str, n: int = 10):
    prompt = f"""
You are an expert in computer vision datasets. 
I need you to generate {n} useful visual descriptions that can help distinguish the category "{category}" in photos.

Requirements:
- The descriptions should be visual features observable in real photos (not abstract concepts).
- Start with very broad, general features, and then make the features progressively more fine-grained and detailed.
- Each description should be a short phrase (not a full sentence).
- Avoid repeating the same feature with only minor wording changes.
- Output them as a numbered list.

Example (for "apple"):
1. round or oval shape
2. smooth surface
3. red, green, or yellow skin color
4. presence of a stem
5. small dimples at the top or bottom
"""
\end{lstlisting}
\captionsetup{labelformat=empty}
\end{figure*}

\section{Conclusion}
We present \momo, a hierarchical representation matching method for CLIP-based CIL. By enriching the semantic space with LLM-generated descriptors, aligning them with hierarchical visual representations, and introducing an adaptive routing mechanism with projection-based updates, our method effectively enhances discrimination and mitigates catastrophic forgetting. Extensive experiments across multiple benchmarks validate the robustness and effectiveness of our method.

\noindent\textbf{Limitations and Future Work.} Our method relies on LLMs to generate textual descriptors for hierarchical representation matching, which may be less effective in scenarios where LLMs fail to generalize. Future directions include leveraging not only the class token but also patch-level tokens to capture richer visual–textual correspondences.

\setcounter{section}{0}
\renewcommand{\thesection}{\Roman{section}}
\makeatletter

\bibliographystyle{IEEEtran}

\bibliography{reference}

\newpage

\end{document}